\documentclass[a4paper,man,floatsintext,apacite]{apa6}

\usepackage[english]{babel}
\usepackage[utf8x]{inputenc}
\usepackage{amsmath}
\usepackage{graphicx}

\usepackage{bm} 
\usepackage{amsfonts} 
\usepackage{float} 
\usepackage{color} 
\usepackage{multirow}

\usepackage[ruled, vlined]{algorithm2e}
\SetKwInput{KwData}{Input}
\SetKwInput{KwResult}{Output}

\graphicspath{{./figs/}}

\title{Optimal Hierarchical Learning Path Design with Reinforcement Learning}
\shorttitle{Optimal Hierarchical Learning Path Design}

\fourauthors{Xiao Li}{Hanchen Xu}{Jinming Zhang
}{Hua-hua Chang}
\fouraffiliations{Department of Educational Psychology\\University of Illinois at Urbana-Champaign}{Department of Electrical and Computer Engineering\\University of Illinois at Urbana-Champaign}{Department of Educational Psychology\\University of Illinois at Urbana-Champaign}{Department of Educational Psychology\\University of Illinois at Urbana-Champaign}

\abstract{E-learning systems are capable of providing more adaptive and efficient learning experiences for students than the traditional classroom setting.
A key component of such systems is the learning strategy, the algorithm that designs the learning paths for students based on information such as the students' current progresses, their skills, learning materials, and etc.
In this paper, we address the problem of finding the optimal learning strategy for an E-learning system.
To this end, we first develop a model for students' hierarchical skills in the E-learning system.
Based on the hierarchical skill model and the classical cognitive diagnosis model, we further develop a framework to model various proficiency levels of hierarchical skills.
The optimal learning strategy on top of the hierarchical structure is found by applying a model-free reinforcement learning method, which does not require information on students' learning transition process. 
The effectiveness of the proposed framework is demonstrated via numerical experiments.}

\keywords{personalized learning, reinforcement learning, hidden Markov model, Markov decision process, cognitive diagnostic model, attribute hierarchy model}

\begin{document}
\maketitle

\section{Introduction}

Designing optimal learning strategies for students has emerged as an interesting and important topic in recent years, along with the trending transformation from traditional classroom teaching to E-learning systems \cite{means2009evaluation}.
Thanks to online learning technologies, information such as students' test results and response time can be monitored, which enables E-learning systems to select the most appropriate learning materials to each individual student.
For example, students are routed with finest learning materials based on their skills, and the materials' contents and difficulty levels, instead of following a routine learning path that does not differentiate individual students.
This notion is referred to as personalized learning \cite{twyman2014competency}, and also known as adaptive learning or smart learning \cite{zhang2016smart}.

Several studies have provided innovative approaches to personalized E-learning systems.
For example, cognitive diagnosis models (CDMs), known as the foundation of assessing students' mastery of skills, are extended to model their learning processes \cite{wang2018tracking,chen2018hidden}. 
The knowledge tracing method \cite{corbett1994knowledge} functions similarly in modeling learning but focusing on one attribute each time \cite{studer2012incorporating}.
In the aforementioned models, skills are assumed to be unstructured without considering skill hierarchical structure and proficiency levels. However, ignoring skill hierarchy and proficiency levels may contaminate classification results \cite{tu2018cognitive}. 
Another direction towards personalized learning is finding optimal learning strategies that recommend learning materials \cite{chen2018recommendation}.
Existing researches typically characterize the learning process as a Markov decision problem, the transition kernel of which is known.
However, the transition kernel is hardly known in practice. 
As a matter of fact, the transition processes of learners' states are unobservable and may vary across different learning materials. 

In this paper, we address those challenges by proposing an integrated E-learning system, which is equipped with the optimal learning strategy obtained via a model-free method that takes the skill hierarchy into account. 
The contributions of this paper are the following.
First, a hierarchical learning model is developed to explicitly characterize skill hierarchy and proficiency levels, which, albeit important, have not been addressed yet in existing models. 
We model the proficiency levels of hierarchical skills following the same form of CDMs; therefore, the latent skills and their proficiency levels can be estimated using CDMs, and the state transitions can be characterized by a hidden Markov model (HMM). 
The proposed hierarchical learning model is easy to implement and can accommodate various types of skill hierarchies \cite{leighton2004attribute}. 
In addition, the number of model parameters and states to be estimated is largely reduced with regards to the restricted state space defined in the model. 
Second, a model-free reinforcement learning (RL) method is applied to finding the optimal learning strategy.
Using RL techniques, the proposed E-learning system is fully data-driven and does not required prior information on the HMM.
At each stage of learning, a set of items will be distributed to the learners, whose responses to these items are next collected by the E-learning system. 
Learners' hidden states are estimated using psychometric models and updated based on the responses. 
We compared the model-free RL method with a heuristic method, and demonstrated via numerical experiments that the model-free RL method can find a better learning strategy that outperforms the heuristic one quickly.

The rest of the paper is organized as follows. 
The CDMs and conventional HMMs used for modeling learning paths are introduced in the ``Preliminaries" section. 
The hierarchical learning model and the model-free RL algorithm for finding the optimal learning strategy is presented in the ``Models and Algorithms" section.
Results from numerical experiments are presented in the ``Experiments" section.
Some concluding remarks and potential future directions are discussed in the ``Concluding Remarks and Future Directions" section.

\section{Preliminaries}

\subsection{Cognitive Diagnosis Models}

CDMs are psychometric models that examine students' mastery of specific skills at a fine-grained level. 
These models provide a summary information in the form of score profiles, the element of which represents the proficiency level of a skill by examinees.
The element takes binary values if only the presence and absence of a skill is modeled.
They are ideal frameworks that aid in identifying optimal learning materials to be distributed next since they keep track of learners' different skills considering their multidimensional features.
Skills in CDMs are discrete and assumed to be latent. 
They are reflected by responses given by examinees to items measuring one or more skills. 
The skill sets are described as \textit{attribute profiles} and each skill is referred to as an \textit{attribute} in the CDMs. 
Binary values are used to model the mastery or non-mastery of a attribute. The proficiency levels of each attribute can be transformed to an attribute profile taking binary values. 
Details of the model will be discussed in the Hierarchical Learning Model section.
As an example, in the deterministic inputs, noisy ``and" gate (DINA) model \cite{junker2001cognitive}---a commonly-used CDM which is both tractable and interpretable, attribute profiles as well as model parameters can be easily estimated by expectation-maximization and Markov Chain Monte Carlo (MCMC) algorithms \cite{de2009dina}.

Most CDMs require the construction of a Q-matrix \cite{embretson1984general} for implementation. 
To be specific, suppose the E-learning system considers $N$ attributes and  contains $J$ items. The Q-matrix is a $J \times N$ matrix whose element $q_{jn}$, $j=1, \cdots, J$, $n=1, \cdots, N$, on the $j^{\text{th}}$ row and $n^{\text{th}}$ column taking binary values, indicates whether the $j^{\text{th}}$ item is associated with the $n^{\text{th}}$ attribute. The DINA model translates the association by a strict rule---the associated attributes are required for learners to answer the item correctly. The Q-matrix specifies the cognitive specification for each test item explicitly \cite{de2009dina}.

An example is provided to illustrate the construction of Q-matrix. Consider the mixed attributes in the system including addition and multiplication. The item ``$5+4$" requires addition attribute to be answered correctly the item, while ``$5+2\times2$" measures both addition and multiplication attributes. Thus the corresponding row of the Q-matrix for the first item is $(1, 0)$ and that for the second is $(1, 1)$. The Q-matrix provides a method to formulate the conditional independence between item responses and attribute profiles. That is, conditioning on measured attributes, item responses are independent of irrelevant attributes. The Q-matrix is generally specified before a test and further improved based on students' responses during the test \cite{liu2012data,chen2018bayesian}.

In the DINA model, the probability of correctly answering an item is defined based on the Q-matrix.
Following the same notation as above, assume $N$ attributes and $J$ items in the E-learning system.
Let $\bm{\alpha_i}$ be the attribute profile for the $i^{\text{th}}$ learner, 
where $\bm{\alpha}_i = (\alpha_{i1}, \alpha_{i2}, \cdots, \alpha_{iN})$ and each element of $\bm{\alpha}_i$ belongs to $\{0, 1\}$. A value of 1 indicates a mastered attribute and 0 an unmastered attribute. 
Let $X_{ij}$ be the response of learner $i$ to item $j$, $j=1, \cdots, J$, where $X_{ij}=1$ indicates a correct answer while $0$ indicates an incorrect one.
Therefore, the probability of a correct answer conditional on the attribute profile is defined as
\begin{equation} \label{eq1}
\mathbb{P}(X_{ij}=1|\bm{\alpha}_i)=(1-s_j)^{\eta_{ij}}g_j^{1-\eta_{ij}},
\end{equation}
where $\mathbb{P}$ denotes probability, $\eta_{ij}$ indicates whether or not the learner $i$ has mastered all attributes required for the item $j$. The value of $\eta_{ij}$ is $1$ if the learner possesses all attributes and is $0$ if the learner lacks at least one of the required attributes. Mathematically, it is defined as
\begin{equation*} \label{eq2}
\eta_{ij}=\prod_{n=1}^N \alpha_{in}^{q_{jn}},
\end{equation*}
$s_j$ denotes the slipping parameter---the probability of a learner possessing all attributes required in item $j$, i.e., $$s_j = \mathbb{P}(X_{ij}=0|\eta_{ij}=1),$$
and $g_j$ denotes the guessing parameter---the probability of correctly answering the item without required attributes, i.e., $$g_j = \mathbb{P}(X_{ij}=1|\eta_{ij}=0).$$

CDMs are classified into non-compensatory and compensatory model \cite{dibello2007review}. 
The DINA model is a non-compensatory model for the reason that it assumes the learner who lacks any of the required attributes will fail to answer the item. Unlike non-compensatory models, compensatory models allow a high ability attribute to compensate for a low ability attribute on another dimension. 
Other non-compensatory models include noisy input, deterministic, ``and" gate (NIDA) model \cite{maris1999estimating}, and the reduced reparameterized unified model \cite{roussos2007skills}.
Compensatory models include deterministic input noisy ``or" gate (DINO) model \cite{templin2006measurement}.
More general CDMs have been developed to include many non-compensatory and compensatory models \cite{henson2009defining,de2011generalized}.
Both non-compensatory and non-compensatory models are well-examined in modeling diagnostic skills.

\subsection{Learning Paths with the Hidden Markov Model}

\begin{figure}[!t]
\centering
\includegraphics[scale=0.5]{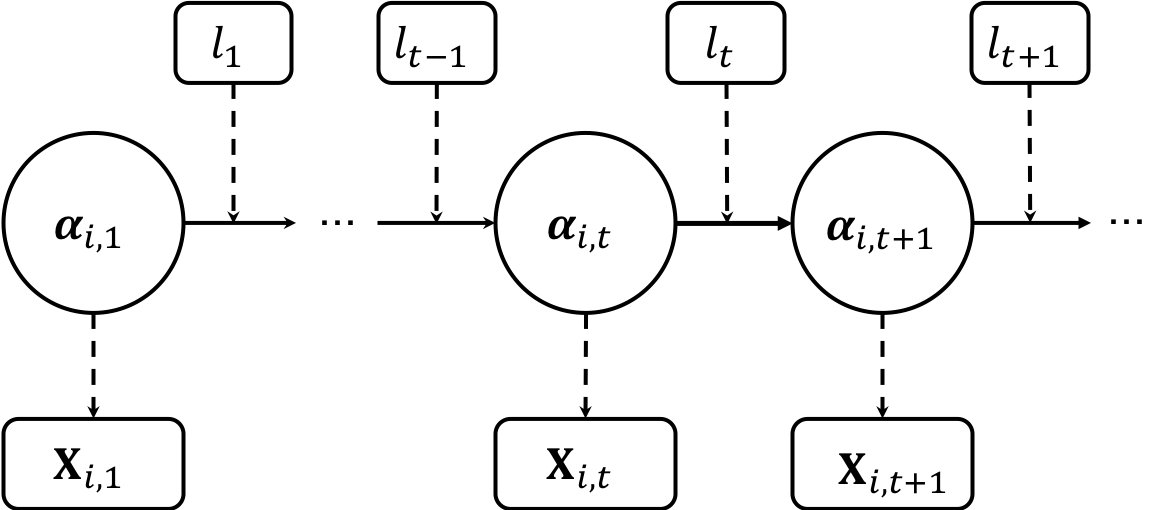}
\caption{An illustration of learning path with the Hidden Markov Model.\label{F7}}
\end{figure}

Learning paths can be modeled by the HMM as the attribute profile is latent \cite{norris1998markov,wang2018tracking}.
The Markov model specifies that a learner's next state, after provided with a certain learning material, will only depend on his or her current state and the material.
Figure \ref{F7} illustrates how to model the learning path with a HMM. Define the attribute profile as the state in the Markov model, denoted as $\bm{\alpha}_{i,t}$ for the $i^{\text{th}}$ learner at time step $t$. The state transition is as follows:
\begin{equation} \label{eq3}
\bm{\alpha}_{i, t} \times l_t \rightarrow \bm{\alpha}_{i, t+1},
\end{equation}
where $l_t$ denotes the learning material distributed at time $t$, and $l_t \in \mathcal{L} = \{l_1, \cdots, l_L\}$, which is the set of all learning materials. 
The transition process from current state to the next is thus formulated as a Markov decision process (MDP). 

The learning paths with latent attribute profiles can either be considered as a partially observable MDP \cite{kaelbling1998planning}, or two separate components, one with a psychometric model and one MDP. In both cases, we assume no retrogress exists---once learners master the attribute, they will not lose it, that is, 
\begin{equation} \label{eq4}
	\mathbb{P}(\bm{\alpha}_{i, t+1} = 1|\bm{\alpha}_{i, t}=1)=1,
\end{equation}
and 
\begin{equation} \label{eq5}
	\mathbb{P}(\bm{\alpha}_{i, t+1} = 0|\bm{\alpha}_{i, t}=1)=0.
\end{equation}

In this study, the psychometric model and a HMM are used to estimate the attribute profiles. Specifically, given time-invariant item parameters and a proper psychometric model such as CDMs, the attribute profile $\bm{\alpha}_{i, t}$ of learner $i$ at time step $t$ can be estimated from item responses. 
Take the DINA model as an example. 
Given item responses from learners at time step $t$, denoted as $\textbf{X}_{i, t}$, the attribute profile can be estimated through \eqref{eq1}. 

\section{Models and Algorithms}

\subsection{Hierarchical Learning Model}
Attribute hierarchy method (AHM) were first proposed to deal with situations where cognitive attributes are hierarchically related and thus dependent \cite{leighton2004attribute}. 
In particular, the AHM investigates precedence ordering of cognitive competencies required to solve test problems. It has four different structures including linear, convergent, divergent and unstructured.
An intuitive example of the hierarchical structure is how students learn addition ``$+$" and multiplication ``$\times$". Addition is considered as a prerequisite for multiplication. Students are able to learn multiplication only after they fully understand addition or at least are equipped with basic knowledge of it. 

\begin{figure}[!t]
\centering
\includegraphics[scale=0.45]{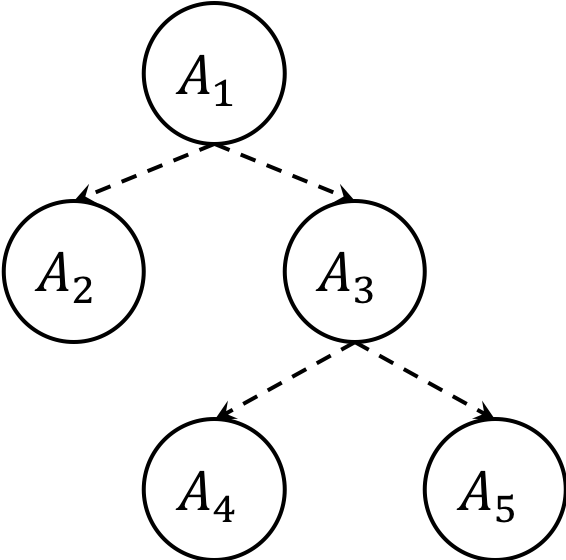}
\caption{A divergent hierarchical structure among five cognitive attributes.\label{F8}}
\end{figure}

All structures investigated by AHM can be split into dependent relationships between two attributes. For example, Fig. \ref{F8} exhibits the divergent structure among 5 attributes, denoted as $A_n$, $n = 1, \cdots, 5$. The hierarchical structure among the five can be split to the four dependent links shown as dotted arrow line in Figure \ref{F8}. That is, $A_1$ is a prerequisite of $A_2$ and $A_3$, while $A_3$ is a prerequisite of $A_4$ and $A_5$. 
Therefore, in order to model the hierarchical structure, we make three assumptions on the link between two dependent attributes.  

Assume attribute $A_1$ is prerequisite to attribute $A_2$. There are $K$ different proficiency levels for each attribute. 
Denote the lack of attribute $A_n$ as $A_n^{(0)}$, $n\in\{1,2\}$, and $K$ different proficiency levels as $A_n^{(1)}, \cdots, A_n^{(K)}$.
Whether or not possessing a certain proficiency level of each attribute is binary. 
We make the following assumptions on the attribute hierarchy:
\begin{enumerate}
\item Learners can only possess a high proficiency level after they have mastered lower proficiency level of the same attribute. That is,
\begin{equation}
\mathbb{P}(A_n^{(k)}=1|A_n^{(k-1)}=0)=0, k \in \{2, \cdots, K\}.
\end{equation}


\item Certain proficiency level of $A_2$ can only be learned after the same proficiency level of $A_1$ is achieved. That is,
\begin{equation}
\mathbb{P}(A_2^{(k)}=1|A_1^{(k)}=0)=0, k \in \{1, \cdots, K\}.
\end{equation}

\item
The probability of a learner to master the attribute $A_{n_1}$ conditional on mastering a high proficiency level of $A_{n_2}$ is no smaller than mastering a lower proficiency level of$A_{n_2}$, $\{n_1, n_2\}=\{1, 2\}$. That is, for $\tilde{k} \in \{1, \cdots, k-1\}$ and $k \in \{2, \cdots, K\}$,
\begin{equation}
\mathbb{P}(A_2^{(k)}=1|A_1^{(\tilde{k}+1)}=1) \geq P(A_2^{(k)}=1|A_1^{(\tilde{k})}=1).
\end{equation}
and for $\tilde{k} \in \{2, \cdots, k\}$ and $k \in \{2, \cdots, K\}$,
\begin{equation}
\mathbb{P}(A_1^{(k)}=1|A_2^{(\tilde{k})}=1) \geq P(A_1^{(k)}=1|A_2^{(\tilde{k}-1)}=1).
\end{equation}
\end{enumerate}
Therefore, by expressing the relationship between dependent attributes, hierarchical attribute structure is modeled. 

We next model different proficiency levels of attributes to be elements of attribute profiles as in CDMs. The proficiency levels of learners on different attributes can be estimated by psychometric models as a result. An example of a Q-matrix for two hierarchical attributes with two proficiency levels is provided in Table \ref{T1}. In this example, attribute addition ($+$) is presumed to be a prerequisite of attribute multiplication ($\times$). One-digit calculation is assumed to be the low proficiency level while two-digit calculation is assumed to be the high proficiency level for both operations. 

\begin{table}[!t]
\centering
\caption{A Q-matrix of Addition ($+$) and Multiplication ($\times$) Attributes with Two Levels. \label{T1}}
\begin{tabular}{lcccc}
\toprule
Item & ${+}^{(1)}$ & ${+}^{(2)}$ & ${\times}^{(1)}$ & ${\times}^{(2)}$\\
\midrule
$7+2$ & 1& 0& 0& 0\\
$11+4*5$& 1& 1& 1& 0\\
$12*31$& 1& 1& 1& 1\\
\bottomrule
\end{tabular}
\end{table}

To incorporate the attribute hierarchy, the state space is constructed following the hierarchical learning model assumptions. Originally in CDMs, $2^4=16$ states shall be included in the HMM with respect to $4$ attributes. With hierarchical learning model, the state space is reduced to $6$ states shown as rows in Table \ref{T2}. As a result, the attribute profile of learner $i$ at time step $t$, i.e., $\bm{\alpha}_{i, t}$, could be any row in Table \ref{T2}.
\begin{table}[!t]
\centering
\caption{State Space for Addition ($+$) and Multiplication ($\times$) Attributes with Two Levels. \label{T2}}
\begin{tabular}{ccccc}
\toprule
State & ${+}^{(1)}$ & ${+}^{(2)}$ & ${\times}^{(1)}$ & ${\times}^{(2)}$\\
\midrule
\texttt{1}& 0& 0& 0& 0 \\
\texttt{2}& 1& 0& 0& 0 \\
\texttt{3}& 1& 1& 0& 0 \\
\texttt{4}& 1& 0& 1& 0 \\
\texttt{5}& 1& 1& 1& 0 \\
\texttt{6}& 1& 1& 1& 1 \\
\bottomrule
\end{tabular}
\end{table}

All attribute hierarchy can be generalized by the hierarchical learning model other than the linear structure given above. More strict assumptions can be added if necessary in practice. For example, a attribute cannot be learned before its prerequisite is fully mastered. The state space of the example in the experiment will be further restricted to the space shown as \text{Table} \ref{T5} with $5$ states only.
\begin{table}[!t]
\centering
\caption{More Restricted State Space for Addition ($+$) and Multiplication ($\times$) Attributes with Two Levels. \label{T5}}
\begin{tabular}{ccccccc}
\toprule
State & ${+}^{(1)}$ & ${+}^{(2)}$ & ${\times}^{(1)}$ & ${\times}^{(2)}$\\
\midrule
\texttt{1}& 0& 0& 0& 0 \\
\texttt{2}& 1& 0& 0& 0 \\
\texttt{3}& 1& 1& 0& 0 \\
\texttt{4}& 1& 1& 1& 0 \\
\texttt{5}& 1& 1& 1& 1 \\
\bottomrule
\end{tabular}
\end{table}

The design of hierarchical learning model makes it possible to incorporate not only attribute hierarchy, but also different  proficiency levels of attributes in CDMs. The model follows the common form of CDMs so that the restricted Q-matrix is easy to construct, and parameters in CDMs as well as attributes can be estimated easily \cite{tu2018cognitive}.
In addition, the hierarchical design largely reduces the number of parameters and attributes to be recovered in CDMs.

\subsection{Reinforcement learning}
RL is widely used in solving problems by interacting with the environment, without requiring an explicitly expressed MDP model \cite{sutton2011reinforcement}.
The RL method can be applied in finding the optimal learning strategy for several reasons.
First, in E-learning systems, how learners' attribute profiles transit after feeding a learning material is unknown. 
RL methods can be an ideal fit in finding the best solution since it does not require an explicit model to estimate the utility of taking actions in the environment \cite{kaelbling1996reinforcement}. 
Second, the learning path with attribute hierarchy modeled by a HMM can be well-solved by the RL method. 
Third, the RL method searches for the long-term optimal solution which takes future rewards into consideration instead of simply choosing the best option at immediate step \cite{littman1994markov}. 
These advantages make it an ideal solution for finding the optimal learning strategy in the E-learning system.

\begin{figure}[!t]
\centering
\includegraphics[scale=0.6]{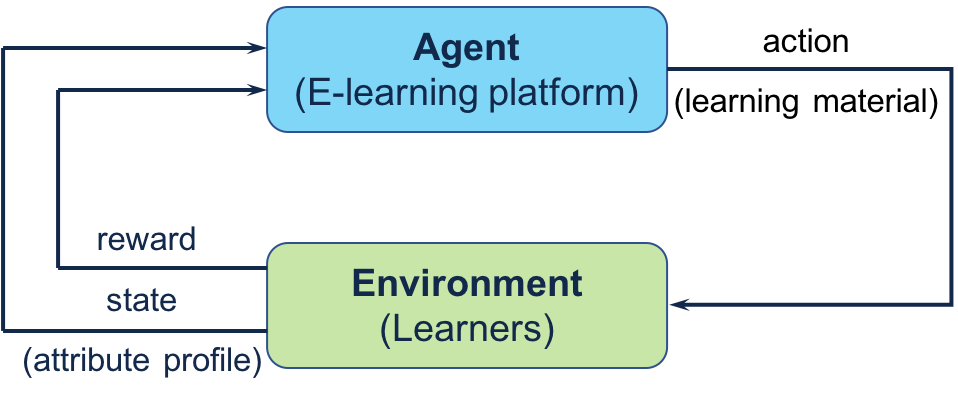}
\caption{RL system in the optimal learning strategy problem.}
\label{F1}
\end{figure}

The overall framework is illustrated in Fig. \ref{F1}, where the agent is the E-learning system that determines action (i.e., learning material), sent to the environment (i.e., learners), which will then send state (i.e., attribute profiles), and a reward signal back to the agent. 

We next model the learning paths as a MDP.
The state space is the set of all attribute profiles $\{\bm{\alpha}\}$.
The action space is defined to be the set of all learning materials $\mathcal{L} = \{l_1, \cdots, l_L\}$.
The reward is shown in Algorithm 1, designed to be decreased if the episode length is too long.
As discussed earlier, the transition kernel satisfies the Markov property.

\begin{algorithm}[!t]
    \SetAlgoLined
    \DontPrintSemicolon
    \KwData{attribute profile (state) set $\{\bm{\alpha}\}$, action set $\mathcal{L}$, learning rate $\beta$, discount factor $\gamma$, decay rate for learning $\lambda_l$, decay rate for exploration $\lambda_e$, initial exploration probability $\epsilon$
    }
    \KwResult{Q function}
	    Randomly initialize the value of $Q(\bm{\alpha}_{i,0}, l_0)$\;
        Receive initial state $\bm{\alpha}_{i,0}$\;
	    \For{$t=0, 1, \cdots$}{
	    	Select $l_t \leftarrow \text{argmax}_lQ(\bm{\alpha}_{i,t}, l)$ with probability of $1-\epsilon$\ and otherwise randomly select $l_t$ with probability of $\epsilon$\;
	    	Receive a new state $\bm{\alpha}_{i,t+1}$\;
	    	Calculate $n_{\bm{\alpha}_{i, t}}$ as the number of mastered attributes at time step $t$ for learner $i$\;
	        Compute reward $r_t$ according to
	        \begin{align*}
	        	r_t = 	
	        	\begin{cases}
					2 (n_{\bm{\alpha}_{i, t+1}}-n_{\bm{\alpha}_{i, t}}) - 0.1 t, & \text{if } n_{\bm{\alpha}_{i, t+1}} > n_{\bm{\alpha}_{i, t}} \\
					-(1+n_{\bm{\alpha}_{i, t}}-n_{\bm{\alpha}_{i, t+1}}) - 0.1 t, & \text{if } n_{\bm{\alpha}_{i, t+1}} \leq n_{\bm{\alpha}_{i, t}}
				\end{cases}
	        \end{align*}\;\vspace{-0.5in}
	    	Calculate $Q$ value
			\begin{align*}
				Q(\bm{\alpha}_{i,t+1}, l_t) := Q(\bm{\alpha}_{i,t}, l_t) + \beta[r_t + \gamma\max_{l'}Q(\bm{\alpha}_{i,t+1}, l') - Q(\bm{\alpha}_{i,t}, l_t)]
			\end{align*}\;\vspace{-0.5in}
			Update learning rate $\beta \leftarrow \beta*\lambda_l$ and exploration rate $\epsilon \leftarrow \epsilon*\lambda_e$\;
	    }
\caption{Q-learning Algorithm for Hierarchical Learning Model}
\label{algo}
\end{algorithm}

Since both the state space and the action space are discrete, a classical model-free RL algorithm---the Q-learning algorithm---can be applied to learn the optimal policy \cite{watkins1992q}.
The Q-learning algorithm estimates an action value function---the so called Q-function---that gives the long-term value of a state-action pair, denoted by $Q(\bm{\alpha}, l)$. 
By taking a discount factor into consideration, the algorithm discounts the future rewards into current time step.
The Q-learning algorithm proves to converge with probability 1 if the learning rate is properly chosen and the state-action space is sufficiently explored \cite{watkins1992q}.
In practice, $\epsilon$-greedy exploration policy is commonly used  with a probability of $\epsilon$ to explore at the beginning and decayed later for exploitation. The detailed algorithm for optimal learning strategy is presented in Algorithm 1.

\section{Experiments}
\subsection{Overview}
The experiment considers two attributes with linear hierarchical structure and three proficiency levels for each attribute. 
Denote the two attributes as $A_1$ and $A_2$. 
The three proficiency levels are represented as $A_1^{(1)}$, $A_1^{(2)}$, $A_1^{(3)}$, $A_2^{(1)}$, $A_2^{(2)}$, and $A_2^{(3)}$ respectively. $A_1^{(0)}$ or $A_2^{(0)}$ is used when the corresponding attribute is not mastered.

Assume $A_1$ is a prerequisite attribute of $A_2$, satisfying all assumptions in the section ``Hierarchical Learning Model".
An intuitive way to understand the hierarchy structure here is to assume $A_1$ to be the addition and $A_2$ to be the multiplication.
The three proficiency levels can be translated to beginner, intermediate and advanced level, while $A_1^{(0)}$ indicates the learner has no knowledge of $A_1$ and so does $A_2^{(0)}$. 

\begin{figure}[!t]
\centering
\includegraphics[width=0.5\textwidth]{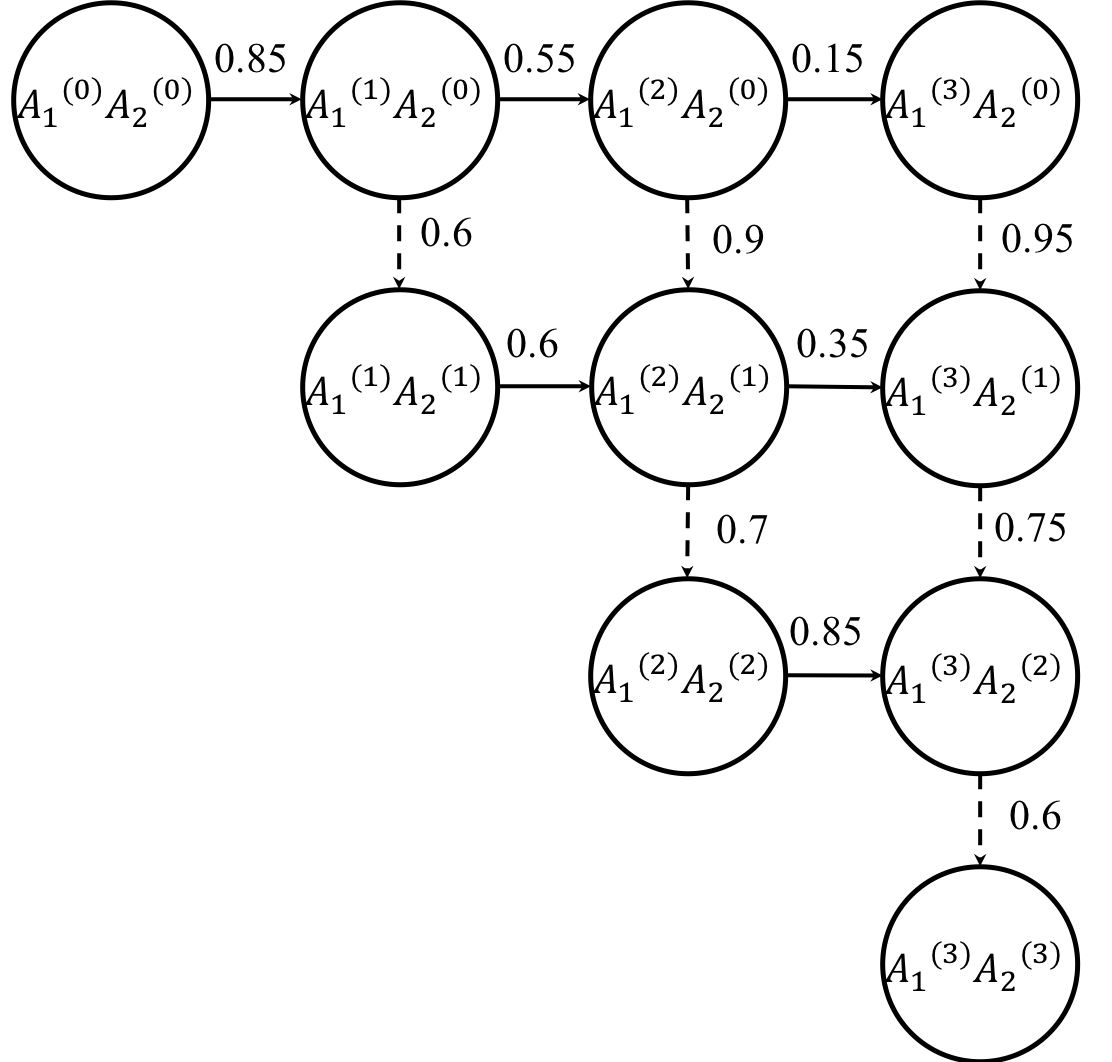}
\caption{The directed graph of the Markov process for the attribute profile consisting of attribute $A_1$ and $A_2$.\label{F2}}
\end{figure}

Assume six learning materials are available, three of which are beginner, intermediate and advanced level materials for attribute $A_1$ and the other three are for attribute $A_2$.
We thus construct the Markov process shown as a directed graph in Fig. \ref{F2}. Each circle represents a state. 
A full arrow shows a transition of attribute $A_1$ while a dotted arrow shows a transition of attribute $A_2$. 
Only one attribute can be improved in each learning step. 
The process satisfies the three assumptions in the hierarchy learning model. Note that the transition from a state to itself is neglected in the directed graph and can be easily calculated by Markov properties. 
The transition matrix, which is unknown to the environment and only applied to predict learners' next state, is constructed accordingly. 

The state space is shown in Table \ref{T4}. 
If the learner acquires the attribute profile of $A_1^{(3)}A_2^{(3)}$, no more learning material will be provided and the learning process ends.
After a learning material is selected by the E-learning system and fed to the learner, a set of test items will be given, to test the learner's current attribute profile. Therefore, the new state can be estimated and updated. 

\begin{table}[!t]
\centering
\caption{State Space for Two Attributes with Three Levels. \label{T4}}
\begin{tabular}{ccccccc}
\toprule
State & $A_1^{(1)}$ & $A_1^{(2)}$ & $A_1^{(3)}$ & $A_2^{(1)}$ & $A_2^{(2)}$ & $A_2^{(3)}$\\
\midrule
\texttt{1}& 0& 0& 0& 0& 0& 0 \\
\texttt{2}& 1& 0& 0& 0& 0& 0 \\
\texttt{3}& 1& 1& 0& 0& 0& 0 \\
\texttt{4}& 1& 1& 1& 0& 0& 0 \\
\texttt{5}& 1& 0& 0& 1& 0& 0 \\
\texttt{6}& 1& 1& 0& 1& 0& 0 \\
\texttt{7}& 1& 1& 1& 1& 0& 0 \\
\texttt{8}& 1& 1& 0& 1& 1& 0 \\
\texttt{9}& 1& 1& 1& 1& 1& 0 \\
\texttt{10}& 1& 1& 1& 1& 1& 1 \\
\bottomrule
\end{tabular}
\end{table}

Figure \ref{F2} reveals the difference between the strategy that only considers immediate reward and the strategy given by RL method that takes future rewards into consideration. 
For instance, suppose a learner reaches the beginner level of the first attribute $A_1$ and has no knowledge of the second attribute $A_2$, i.e., in state $A_1^{(1)}A_2^{(0)}$. 
The beginner level material for attribute $A_2$ gives the shortest expected learning time at this step defined as $\sum_{t=1}^{\infty}t\mathbb{P}(\alpha_{s,t}=1|\alpha_{s,0}=0)$ which is $1/0.6 \approx 1.67$. 
However, although the intermediate level material for attribute $A_1$ brings relatively longer learning time, leading to less rewards at current step, the overall expected learning time of path through $A_1^{(2)}A_2^{(0)}$ to $A_1^{(2)}A_2^{(1)}$, which is $1/0.55 + 1/0.9\approx 2.93$, is less than that through $A_1^{(1)}A_2^{(1)}$ to $A_1^{(2)}A_2^{(1)}$, which is $1/0.6 + 1/0.6\approx 3.33$. 
As a result, although to learn beginner level attribute $A_2$ first is quicker at current step, it is not the most optimal learning strategy overall.

In order to simulate the psychometric model estimation step, an estimation error of $0.05$ was added to the state, indicating there is a $5\%$ probability that the estimated state is incorrect. 
In CDM researches, the average pattern correct classification rate (PCCR) is usually larger than $95\%$. 
Therefore, an estimation error of $0.05$ is large enough to show the reliability of the optimal learning strategy.
In addition, simulation results for cases with an estimation error ranging from $1\%$ to $10\%$ are included to show that the Q-learning algorithm is reliable and stable to find the optimal learning strategy even with the presence of estimation error. 
In practice, the states are estimated and updated from responses of test items and item parameters by psychometric models.

The rest of parameters are as follows: initial learning rate $\beta = 0.01$, discount factor $\gamma = 0.99$, and initial exploration probability $\epsilon = 1$. 
A decay rate of $0.999$ is applied for $\beta$ and a decay rate of $0.99$ is used for $\epsilon$. 
Therefore, after $5000$ episodes, the learning rate $\beta$ decays to a value of $0.7\%$ and the exploration probability $\epsilon$ decays to $1.50 \times 10^{-22}$.

The Q-learning algorithm is trained in $5000$ episodes. 
After that, the trained model is applied in another $1000$ episodes and compared with a heuristic strategy, which selects the next learning material that can improve the learner's proficiency level in accordance with hierarchical learning model assumptions. 
For instance, if the learner's attribute profile is estimated to be $A_1^{(1)}A_2^{(0)}$, the learning material will be selected from beginner level material for attribute $A_2$ and intermediate level material for attribute $A_1$. 
The two methods are compared under both with and without estimation error.

Two numerical experiments are conducted in the study. 
In the first experiment, the initial states for all learners are $A_1^{(0)}A_2^{(0)}$, which means none of the learners have any knowledge of the two attributes. 
In the second experiment, learners start with different proficiency levels except for $A_1^{(3)} A_2^{(3)}$. 
The second experiment shows that as long as the learner has not fully mastered attributes in the E-learning system, no matter where they begin with, the system can find the optimal learning strategy for each of them.

\subsection{Results}

\subsubsection{Learning Strategy Comparison}

\begin{figure}[!t]
\centering
\includegraphics[scale=0.7]{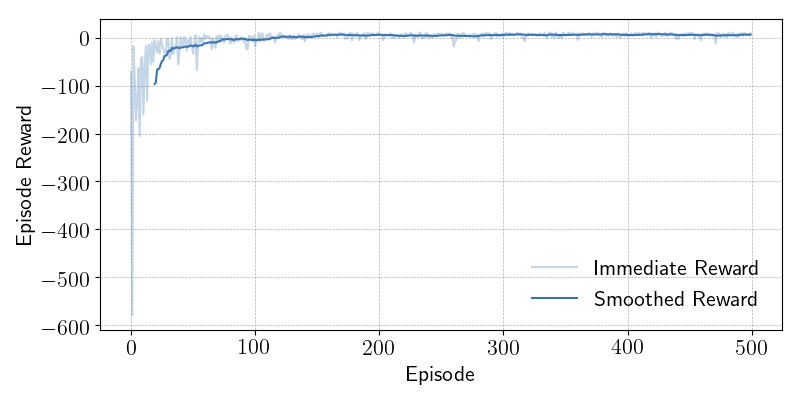}
\caption{Rewards under optimal learning strategy without estimation error.\label{F3}}
\end{figure}

\begin{figure}[!t]
\centering
\includegraphics[scale=0.7]{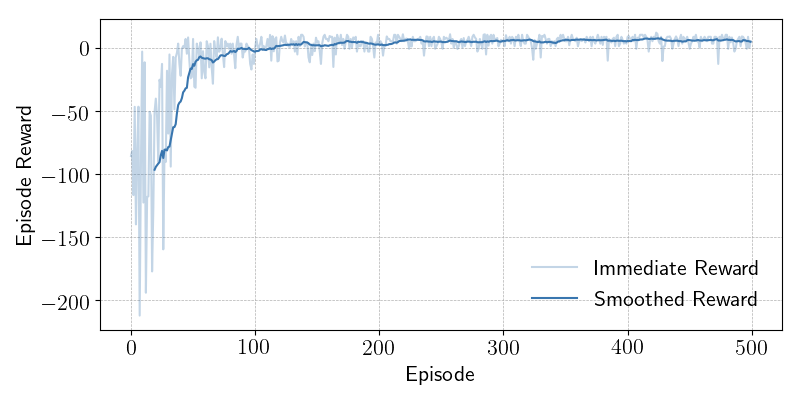}
\caption{Rewards under optimal learning strategy with estimation error.\label{F4}}
\end{figure}

Figures \ref{F3} and \ref{F4} present the rewards under the RL method across $1000$ episodes, including both the immediate reward and the smoothed reward with a smoothing window of $20$.
Figure \ref{F3} shows that the reward becomes stable after $200$ episodes under RL method without estimation error, which means the method finds the optimal strategy after training on $200$ students. The result indicates that the RL method finds the optimal learning strategy quickly. After a $5\%$ estimation error is added to the system, the Figure \ref{F4} presents that the RL method still finds the optimal learning strategy after around $250$ episodes. 

\begin{figure}[!t]
\centering
\includegraphics[scale=0.7]{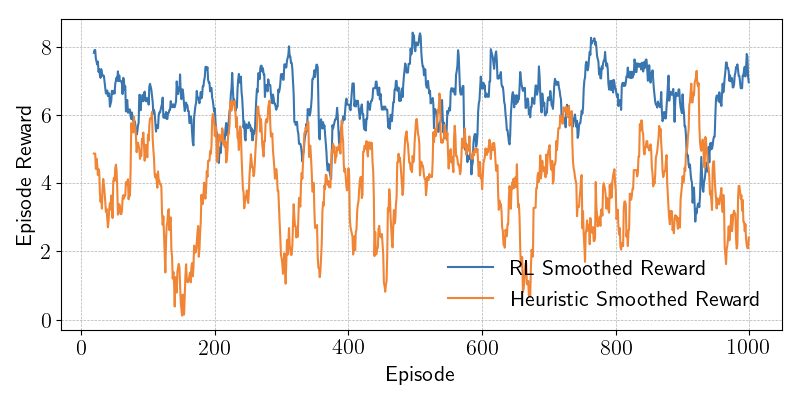}
\caption{Smoothed rewards under optimal strategy learned via RL and heuristic strategy without estimation error.\label{F5}}
\end{figure}

\begin{figure}[!t]
\centering
\includegraphics[scale=0.7]{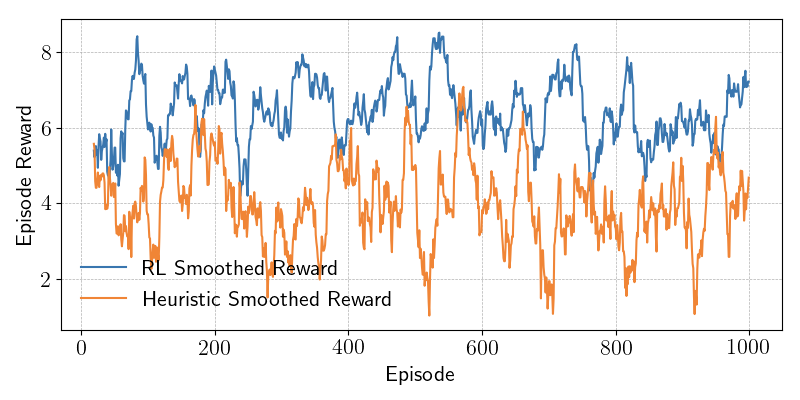}
\caption{Smoothed rewards under optimal strategy learned via RL and heuristic strategy with estimation error.\label{F6}}
\end{figure}

Figures \ref{F5} and \ref{F6} give a comparison between the RL method and heuristic method across $1000$ episodes where the RL method has been trained in $5000$ episodes and applied to new students. 
No estimation error is added in Fig. \ref{F5} while a $5\%$ estimation error is added to both methods in Fig. \ref{F6}. Both figures show that the reward under the RL method is higher than the heuristic method. The smoothed reward of the RL method is significantly higher than that of the heuristic method in both with or without estimation error. 

\text{Table} \ref{T3} shows the overall mean and standard deviation of rewards and episode lengths in two methods. The RL method has much higher mean and lower standard deviation of rewards than the heuristic method, together with shorter episode lengths and smaller episode length standard deviation as well.
It is worth noting that although the average episode length with $5\%$ estimation error is slightly higher than that without estimation error, the difference is minimal.

\begin{table}[!t]
	\renewcommand{\arraystretch}{1.2}
	\caption{Mean and Standard Deviation (SD) of Rewards and Episode Lengths (EL). \label{T3}}
	\label{table:linear}
	\centering
	\begin{tabular}{clcc}
		\toprule
		& Methods & RL & Heuristic\\
		\midrule
		\multirow{4}*{\rotatebox{90}{No Estimation Error}}
		& Reward mean & 6.43 & 3.99 \\
		& Reward SD & 3.61 & 5.34 \\
		& EL mean & 7.34 & 8.57 \\
		& EL SD & 1.90 & 2.62 \\
		\midrule
		\multirow{4}*{\rotatebox{90}{$5\%$ Estimation Error}}
		& Reward mean & 6.41 & 3.98 \\
		& Reward SD & 3.60 & 5.37 \\
		& EL mean & 7.73 & 9.01 \\
		& EL SD & 2.07 & 2.74 \\
		\bottomrule
	\end{tabular}
\end{table}

\begin{figure}[!t]
\centering
\includegraphics[scale=0.7]{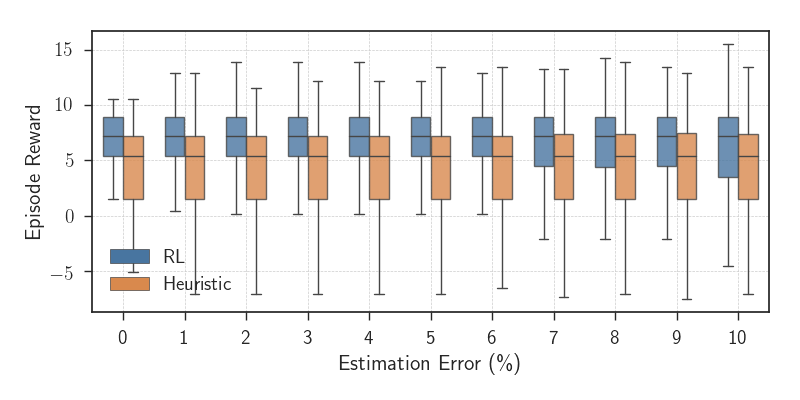}k
\caption{Comparison of rewards under optimal strategy learned via RL and heuristic strategy with estimation errors.\label{F10}}
\end{figure}

Figure \ref{F10} gives a comparison between the RL method and heuristic method across $1000$ episode under $10$ different estimation errors and no estimation error using the box plot where the RL method has been trained in $5000$ episodes. 
The figure shows that the average award under the RL method is much higher than that under the heuristic method across $11$ estimation errors. 
In addition, the RL method also produces smaller standard deviation of rewards than the heuristic method. 
Although the standard deviation of the RL method tends to increase when the estimation error increases, it is still smaller than that of the heuristic method.  

The simulation results shown above indicate that the RL method finds the better learning strategy than heuristic method. More importantly, the estimation error has negligible impact on the performance of RL method in searching for optimal strategy.

\subsubsection{Impacts of Various Initial States}

\begin{figure}[!t]
\centering
\includegraphics[scale=0.7]{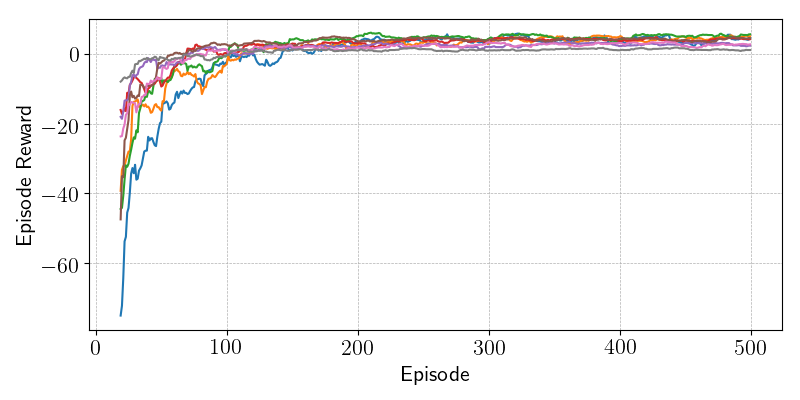}
\caption{Smoothed rewards of different initial states under optimal learning strategy with $5\%$ estimation error.\label{F9}}
\end{figure}

Figure \ref{F9} presents the smoothed rewards of nine different initial states other than $A_1^{L_0}A_2^{L_0}$, with a smoothing window of $20$. 
A $5\%$ estimation error is added to the system to simulate realistic cases.
The result demonstrates that the RL method can quickly find the optimal learning strategy for all learners with different initial attributes. The algorithm converges after $200$ episodes indicating that the optimal strategy can be found after it is trained on only $200$ learners. Therefore, once a learner's initial attribute is estimated by a set of items, the learner can follow the optimal learning strategy to acquire new attributes with the fastest route provided by the system. 

\section{Concluding Remarks and Future Directions}


In this paper, we proposed a hierarchical learning model that incorporates attribute hierarchy and proficiency levels of attributes together in the E-learning system.
The model follows the same form of discrete attributes and Q-matrix required by CDMs so that parameters and hidden states can be easily recovered and estimated. 
In addition, the transition process for student learning is formulated as a MDP. 
Then, a model-free RL method is applied to finding the optimal learning strategy on top of the hierarchical framework.

Experiment results suggest that the optimal design with the RL method outperforms the heuristic strategy substantially with and without the estimation error. 
The mean and the standard deviation of the learning episode length achieved by the RL method is significantly smaller compared to those obtained in the heuristic method.
In addition, the RL method can find the optimal learning strategy quickly for all learners with different initial attribute proficiency levels.
As a result, learners with various proficiency levels will be fed with the most appropriate material at each step.
To implement the system in the real world, a set of items will be given to learners after they finish each learning stage. 
Their attributes will be estimated and the state can be updated based on their responses to the given items. 

Several directions are possible for future researches. 
First, other dimensionality methods can be applied to classify learners at the first stage \cite{zhang1999theoretical,zhang2013procedure}, in addition to using estimation method to get learners' initial states. It is important to have an accurate estimation learners' initial states so that the most appropriate optimal learning strategy can be distributed to each individual.
Second, different algorithms can be proposed to selects the personalized learning materials that can maximize learners' immediate or future rewards \cite{manickam2017contextual}.
Lastly, learners' attributes are restricted to a state space satisfying hierarchical learning model assumptions. 
CDMs with restricted state space as well as Q-matrix can be further explored \cite{tu2018cognitive}.
The identifiability conditions for the restricted latent structure model shall also be rigorously studied \cite{xu2017identifiability}.

\bibliographystyle{apacite}
\bibliography{ref}

\end{document}